\pdfoutput=1

\documentclass[11pt]{article}

\usepackage{EMNLP2023}

\usepackage{times}
\usepackage{latexsym}
\usepackage{graphicx}
\usepackage{caption}
\usepackage{subcaption}
\usepackage{amsfonts}
\usepackage{inconsolata}
\usepackage{amsmath}
\usepackage{amssymb}
\usepackage{booktabs}
\usepackage{multirow}
\usepackage{algorithm}
\usepackage{algorithmic}
\usepackage{amsthm}
\usepackage{CJKutf8}
\usepackage{array}
\usepackage{multirow}
\usepackage{color}
\usepackage{tcolorbox}
\usepackage{adjustbox}
\tcbset{width=0.9\textwidth,boxrule=0pt,colback=red,arc=0pt,auto outer arc,left=0pt,right=0pt,boxsep=5pt}
\newtheorem{theorem}{Theorem}
\newcolumntype{P}[1]{>{\centering\arraybackslash}p{#1}}
\usepackage[T1]{fontenc}

\usepackage[utf8]{inputenc}

\usepackage{microtype}

\usepackage{inconsolata}

\title{Topic-Aware Causal Intervention for Counterfactual Detection}

\author{Thong Nguyen \\
  National University of Singapore \\
  \texttt{e0998147@u.nus.edu} \\\And
  Truc-My Nguyen \\
  Ho Chi Minh city University of Technology \\
  \texttt{nguyenmy3399@gmail.com} \\}

\begin{document}
\maketitle
\begin{abstract}

Counterfactual statements, which describe events that did not or cannot take place, are beneficial to numerous NLP applications. Hence, we consider the problem of counterfactual detection (CFD) and seek to enhance the CFD models. Previous models are reliant on clue phrases to predict counterfactuality, so they suffer from significant performance drop when clue phrase hints do not exist during testing. Moreover, these models tend to predict non-counterfactuals over counterfactuals. To address these issues, we propose to integrate neural topic model into the CFD model to capture the global semantics of the input statement. We continue to causally intervene the hidden representations of the CFD model to balance the effect of the class labels. Extensive experiments show that our approach outperforms previous state-of-the-art CFD and bias-resolving methods in both the CFD and other bias-sensitive tasks.

\end{abstract}

\section{Introduction}
Counterfactual statements describe an event that may not, did not, or cannot occur, and the consequence(s) that did not occur as well \citep{o2021wish}. For example, consider the statement --- \emph{I would purchase this physics book, but I really want that my brain has a tiny amount of interest in science!}. We can partition the statement into two components: a component about the event (\emph{my brain has a tiny amount of interest in science}) as the antecedent, and the consequence of the event (\emph{I would purchase this physics book}) as the consequent. Both the antecedent and the consequent did not take place (\emph{neither the speaker has purchased the book nor he is interested in science}). Accurate detection of such counterfactual statements is beneficial to various NLP applications, such as in social media or psychology. In social media, counterfactual detection (CFD) can be helpful by eliminating irrelevant content \citep{o2021wish}. For example, in the previous statement, we should not return science or physics content because the user is not interested. Detecting counterfactuality can also give useful features to perform psychology assessment of huge populations \citep{son2017recognizing}.

\renewcommand{\arraystretch}{1.1}{
\begin{table}[t]
\small
\centering
\resizebox{\linewidth}{!}{
\begin{tabular}{p{1.8cm}|p{3.8cm}|p{1.6cm}|c}
\hline
\textbf{Scenarios} & \textbf{Examples} & \textbf{mBERT Predictions} & \textbf{Labels} \\ \hline
\multirow{13}{\linewidth}{Clue phrase Anomaly} & It doesn’t work as well as I was hoping it would, it is a waste of money. & \multirow{3}{\linewidth}{Negative} & \multirow{6}{*}{Positive} \\ \cline{2-3}
& I don’t like to go into the plot a lot. The blurb represents the book fairly. & \multirow{3}{*}{Negative} & \\ \cline{2-4}
 & Who would have thought a pillow could make such a difference. & \multirow{3}{\linewidth}{Positive} & \multirow{7.5}{*}{Negative} \\ \cline{2-3}
& The girlfriend was annoying, and it made me wonder if any man in his right mind would have put up with her behavior as long as he did. & \multirow{5}{*}{Positive} & \\ \hline
\multirow{6.0}{\linewidth}{Cross-lingual input} & It would have been, people would say, worse than Watergate. & \multirow{3}{\linewidth}{Positive} & \multirow{6.0}{*}{Positive} \\ \cline{2-3}
& \begin{CJK}{UTF8}{min}ウォーターゲート事件よりもひどかったかもしれない、と人々は言うだろう。\end{CJK} & \multirow{3}{\linewidth}{Negative} & \\
\hline
\end{tabular}}
\caption{Examples of counterfactual detection from the Amazon-2021 dataset. We denote mBERT predictions of \emph{positive} (counterfactual) and \emph{negative} (non-counterfactual) classes.}
\label{tab:case_study}
\vspace{-20pt}
\end{table}}

Previous development of monolingual and multilingual CFD methods depend on extensive labelled datasets \citep{o2021wish}. However, in CFD datasets, the percentage of counterfactual examples is heavily low, even approaching $1-2\%$ \citep{son2017recognizing}. This class imbalance has two weaknesses. First, because counterfactual hints are so limited for the CFD model to learn, it tends to rely on clue phrases, e.g. \emph{if}, \emph{I wish}, etc., to detect counterfactuality. When the existence of such clue phrases does not correlate with the counterfactuality, the model might be led to false predictions. As illustrated in Table \ref{tab:case_study}, the mBERT baseline predicts all incorrect classes for both counterfactual examples, which do not include clue phrases, and non-counterfactual ones, which include clue phrases. Moreover, the performance might substantially drop if the model is tested upon a language different from the training language. As shown in Table \ref{tab:case_study}, the multilingual mBERT predicts the correct class for the English statement, but misclassifies the Japanese one of similar meaning. Second, the class imbalance causes the CFD model to bias towards the non-counterfactual class over the counterfactual one, thus resulting in sub-optimal performance.

\begin{figure}[t]
\centering
\includegraphics[width=\linewidth] {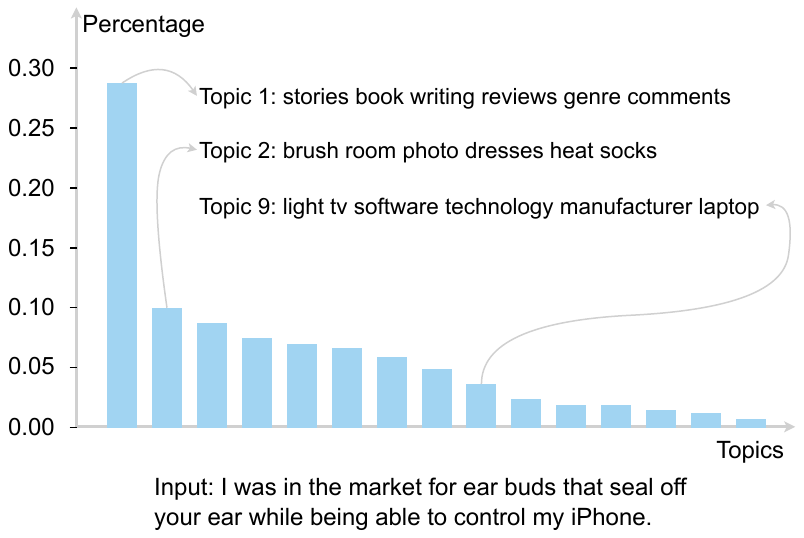}
\caption{For each topic, we count the percentage of inputs in which the topic has the largest probability in the topic representation.  Topic $1$, $2$, and $9$ refer to three top topics of the input document, in descending order of probability.}
\label{fig:topic_distribution}
\vspace{-15pt}
\end{figure}

To address the first issue, we propose to incorporate neural topic model (NTM) into the CFD module. Particularly, we aim to approximate the global semantics of the input statement learned from the posterior distribution of the NTM. The posterior distribution generates the global semantics in terms of the topic representation to guide the CFD model towards semantics of the input instead of the clue phrases. However, a challenge exists that the NTM tends to repetitively assign large weights to a certain small group of topics. In Figure \ref{fig:topic_distribution}, even though the input statement is about \emph{ear buds} and \emph{iPhone}, the NTM still infers it to be highly related to \emph{stories}, \emph{book}, and \emph{reviews}. To cope with this challenge, we propose to adapt backdoor adjustment that adjusts the behavior of neural topic model to make it consider all topics fairly. To the best of our knowledge, no study has explored the benefit of backdoor adjusted NTM for counterfactual detection.

To address the second issue, we view the CFD problem from a causal perspective. Our perspective gives rise to a causal graph where the class imbalance plays a confounder role in influencing hidden representations of the input statement. Based on the graph, we propose to perform causal intervention on these representations to remove the confounding effect of the imbalance phenomenon and enhance the model prediction. 

To sum up, our contributions are as follows:
\begin{itemize}
    \item We propose a novel neural topic model equipped with the backdoor adjustment to produce effective topic representations for benefiting counterfactual detection.
    \item We propose causal intervention upon hidden representations to ameliorate the confounding effect of the class imbalance in counterfactual detection datasets.
    \item Extensive experiments demonstrate that we significantly outperform state-of-the-art CFD and bias-resolving approaches. Our method is also applicable to other bias-sensitive natural language understanding tasks. 
\end{itemize}
\section{Related Work}
\subsection{Representational Intervention for Deep Learning}
Representational intervention has been popularly adopted in deep learning applications. Some include document summarization \cite{nguyen2021enriching, nguyen2022improving}, topic modeling \citep{wu2024fastopic, wu2023infoctm, wu2024affinity, wu2023effective, wu2024modeling}, document ranking \citep{nguyen2023gradient, nguyen2022adaptive}, sentiment analysis \citep{nguyen2023expand, nguyen2024kdmcse, nguyen2023improving}, video moment retrieval \cite{nguyen2023demaformer, nguyen2024read}, and video question answering \cite{nguyen2024video, nguyen2024encoding}. As one notable approach for representational intervention, causal inference has attracted myriad attention as a method to interpret adversarial attacks \cite{zhao2022certified} and eradicate spurious confounding factors in SGD optimizer \cite{tang2020long}.

\subsection{Predictive Biases in Deep Learning}
Research community has long searched for objective-based and augmentation-based countermeasures against biases that drive deep learning models to ignore the input content when making predictions \citep{wu2024survey, nguyen2024meta, nguyen2021contrastive, nguyen2024topic}. For the objective-based direction, \citet{karimi2020end} propose to increase the loss weight of rare examples and subtract the gradients of the biased model from the main one to mitigate their spurious influences. In the second direction, \citet{wang2022causal} perturb words to prevent the confounding effect of language bias. \citet{wang2021robustness} suggest augmenting the original training set with samples containing antonyms of high coefficient terms and reverse label. However, their method demands human supervision and solely involves sentiment classification. Focusing on Counterfactual Detection, \citet{o2021wish} decide to mask clue phrases and populate counterfactual examples through backtranslation. Nevertheless, they find that these methods suffer from deficiency since counterfactuality also depends on the context. Contrast to them, we decide to causally intervene the contextualized representations to reduce the confounding effect of the biases.
\section{Methodology}
In this section, we sequentially formulate the preliminaries of counterfactual detection and neural topic model, introduce our proposed causal perspective for the task, and then articulate the implementation details of our framework.
\subsection{Problem Formulation}
Given an input sentence $S = \{w_1, w_2, …, w_N\}$ and its bag-of-word (BOW) representation $\mathbf{x}_{\text{bow}}$, we aim to train a model function $f$ that maps $S$ and $\mathbf{x}_{\text{bow}}$ to a probability scalar $y \in [0,1]$. The probability magnitude will denote whether the input sentence is counterfactual or not. 

\subsection{Neural Topic Model (NTM)}
Our neural topic model possesses the Variational AutoEncoder architecture \cite{miao2017discovering, kingma2013auto}. It consists of an encoder to produce topic representation and a decoder to reconstruct the original input based upon the representation. 

\noindent\textbf{Topic Encoder.} Its function is to encode the input $\mathbf{x}_{\text{bow}}$ into the topic representation $\boldsymbol{\theta}$. In the beginning, $\mathbf{x}_{\text{bow}}$ is forwarded to both non-linear and linear layers to estimate the mean $\boldsymbol{\mu}$ and standard deviation $\boldsymbol{\sigma}$ of the variational distribution $q(\mathbf{z}|\mathbf{x})$:
\begin{equation}
    \pi = f_0 (\mathbf{x}_{\text{bow}}),\; \boldsymbol{\mu} = f_\mu (\pi), \; \log \boldsymbol{\sigma} = f_\sigma (\pi),
\end{equation}
where we implement $f_0$ as a non-linear layer with the softplus activation function; $f_\mu$ and $f_\sigma$ are two linear layers. Subsequently, to lessen the gradient variance, we adapt the reparameterization trick \cite{kingma2013auto} to draw the latent vector z:
\begin{equation}
\mathbf{z} = \boldsymbol{\mu} + \boldsymbol{\sigma} \cdot \boldsymbol{\epsilon}, \quad \boldsymbol{\epsilon} \sim \mathcal{N} (\mathbf{0}, \mathbf{I}).
\end{equation}
Then, we normalize $\mathbf{z}$ with the softmax function to attain the topic representation $\boldsymbol{\theta}$ as:
\begin{equation}
    \boldsymbol{\theta} = \text{softmax} (\mathbf{z}).
\end{equation}

\noindent\textbf{Topic Decoder.} Given the topic representation $\boldsymbol{\theta}$, the decoder’s task is to reconstruct the original input $\mathbf{x}_\text{bow}$ as $\mathbf{x'}_\text{bow}$. It performs the sampling process to extract the word distribution:
\begin{itemize}
    \item For each word $w \in \mathbf{x}_{\text{bow}}$, draw $w \sim \text{softmax} (f_\phi (\boldsymbol{\theta}))$,
\end{itemize}
where $f_\phi$ denotes a ReLU-activated non-linear transformation. In the ensuing sections, we designate the weight matrix $\phi = (\boldsymbol{\phi}_1, \boldsymbol{\phi}_2, …, \boldsymbol{\phi}_K) \in \mathbb{R}^{V \times K}$ of $f_\phi$ as the topic-word distribution, in which $V$ and $K$ denote the vocabulary size and the number of topics, respectively. We also leverage the topic representation $\boldsymbol{\theta}$ as global semantics to enhance the counterfactual detection module.

\subsection{Causal Perspective into Counterfactual Detection}
\label{subsect:causal_perspective}
\begin{figure}[t]
\centering
\includegraphics[width=\linewidth] {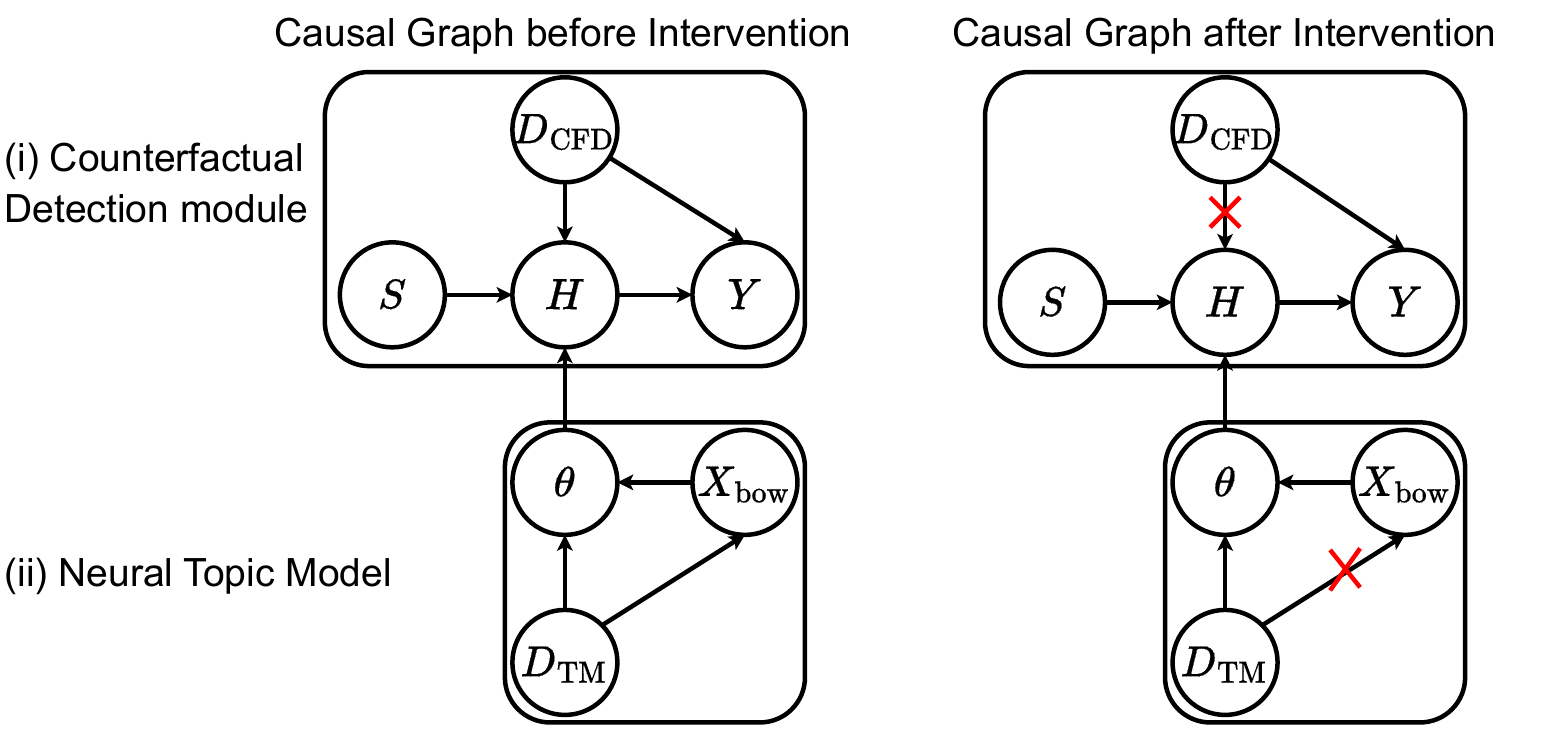}
\caption{(left) Our proposed causal model for counterfactual detection. (right) The causal graph after removing arrows from $D_{\text{CFD}}$ to $H$ and $D_{\text{TM}}$ to $X_{\text{bow}}$, eliminating spurious effects of the label and topic biases.}
\label{fig:causal_graph}
\vspace{-10pt}
\end{figure}
To investigate the relation among factors in our counterfactual detection system, we propose a structural causal graph (SCG) in Figure \ref{fig:causal_graph}. Our graph includes vertices, direct edges, and two sub-graphs for denoting random variables, causal effects, with the pre-intervened and post-intervened states, respectively.

\noindent\textbf{SCG for Topic Modeling.} In this component, the topic bias $D_{\text{TM}}$ is the confounder that influences variables $\theta$ and $X_{\text{bow}}$ in the neural topic model.

\begin{itemize}
\item $X_{\text{bow}} \leftarrow D_{\text{TM}} \rightarrow \theta$: This backdoor path elicits the spurious correlation between $\mathbf{x}_{\text{bow}}$ and $\boldsymbol{\theta}$ instances. In topic modeling, neural topic models have a tendency to align documents with a repetitive set of topics.
\item $\theta \rightarrow H$: Because of the confounder $D_{\text{TM}}$, the inferred global semantics might comprise irrelevant entries that do not represent the true semantics of the document. Therefore, the fallacious semantics could become detrimental noise to adulterate the hidden representations $\mathbf{h}$, which is the direct input to the counterfactual classifier.
\end{itemize}

\noindent\textbf{SCG for Counterfactual Detection.} This component delineates causalities among four variables in counterfactual detection: input sequence $S$, encoded content $H$, output prediction $Y$, and the imbalanced label bias $D_{\text{CFD}}$. In detail, the imbalanced label distribution confounds both the predicted output $Y$ and variable $H$, leading to erroneous correlation between $H$ and $Y$.

\begin{itemize}
\item $H \leftarrow D_{\text{CFD}} \rightarrow Y$ specifies the effect of $D_{\text{CFD}}$ on hidden representations. In practice, the overwhelming population of the negative label in counterfactual datasets might result in learned representations that mostly express non-counterfactual features, thus driving the detection model towards the non-counterfactual response during prediction.
\end{itemize}

\noindent\textbf{Causal Intervention on Textual Representations.} We now present the method to remove the confounding effects. To obtain the deconfounded representations, we capture the causal effect from $X_{\text{bow}}$ to $\theta$ and from $H$ to $Y$ via the Causal Intervention technique, i.e. Backdoor Adjustment \cite{pearl2009causal}, with the following theorem to remove the arrow from $D_{\text{TM}}$ to $X_{\text{bow}}$ and $D_{\text{CFD}}$ to $H$.
\begin{theorem}
 (Backdoor Adjustment \cite{pearl2009causal}) Let $o \in \{y, \boldsymbol{\theta}\}$, $i \in \{\mathbf{x}_{\textup{bow}}, \mathbf{h}\}$, and $n \in \{d_{\textup{TM}}, d_{\textup{CFD}}\}$. Then,
\begin{equation}
 p(o | \textup{do}(i)) = p^{N \nrightarrow I}(o|\textup{do}(i)) = \sum_{n} p(o|i, n) \cdot p(n).
 \label{eq:backdoor_adjustment}
\end{equation}
\end{theorem}
\noindent This theorem shows that we can model the deconfounded likelihood $p(o|\textup{do}(i))$ through estimating $p^{N \rightarrow I}(o|i, n)$ and $p(n)$. We will expound the implementation of $p(o |\textup{do}(i))$ to assist the model in predicting counterfactuality in Section \ref{subsect:model_implementation} and deconfound neural topic model in Section \ref{subsect:training_strategy}.
\subsection{Model Implementation}
\label{subsect:model_implementation}
\begin{figure*}[t]
\centering
\includegraphics[width=0.75\textwidth]{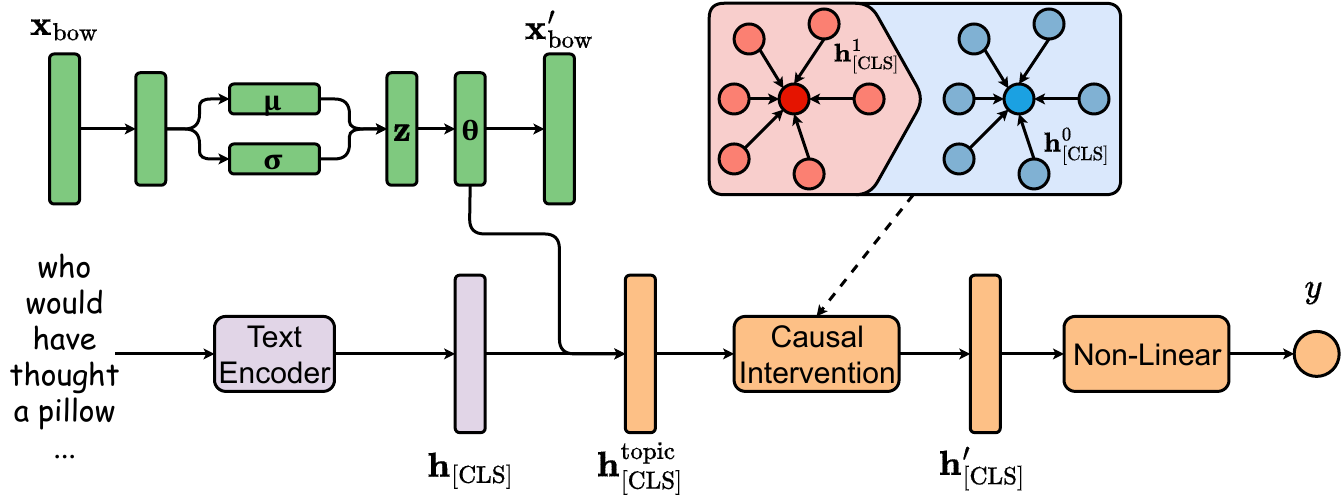}
\caption{Illustration of the Topic-aware Causal Intervention Framework for Counterfactual Detection. Here the green component denotes the neural topic model, the purple component the text encoder, and the orange component our causal intervention operation for counterfactuality prediction.
}
\label{fig:overall_model}
\vspace{-10pt}
\end{figure*}
Our overall framework is illustrated in Figure \ref{fig:overall_model}.

\noindent \textbf{NTM for Text Encoder.} To address the issue of model reliance on clue phrases in counterfactual detection, we propose to condition contextualized representations on global semantics yielded by the neural topic model. 

Initially, we append the special token [\text{CLS}] to the beginning of the input sequence. Then, the text encoder converts each discrete token $w_i$ into the hidden vector $\mathbf{h}_i$ as:
\begin{equation}
\begin{split}
\mathbf{h}_{[\text{CLS}]}, \mathbf{h}_1, …, \mathbf{h}_{|S|} = \text{TextEncoder} (\text{[\text{CLS}]}, \\ w_1, …, w_{|S|}).
\end{split}
\end{equation}
We insert global semantics $\boldsymbol{\theta}$ of the input $S$ into the encoded information: 
\begin{equation}
\mathbf{h}^{\text{topic}}_{i} = \text{tanh}(\text{Linear}([\mathbf{h}_i, \boldsymbol{\theta}]),
\end{equation}
where [,] denotes the concatenation operator.

\noindent \textbf{Causal Intervention for Predicting Counterfactuality.} As mentioned in Section \ref{subsect:causal_perspective}, we propose to debias hidden vectors from the imbalanced label bias. To this end, we set $p(d_{\text{CFD}}) = \frac{1}{|\mathcal{Y}|}$, where $\mathcal{Y}$ is the set of groundtruth labels. Formally, the Eq. (\ref{eq:backdoor_adjustment}) becomes:
\begin{equation}
p(y | \textup{do}(c)) = \frac{1}{|\mathcal{Y}|} \sum_{d_{\text{CFD}}} p(y|c, d_{\text{CFD}}).
\end{equation}
Because of $d_{\text{CFD}}$, we need to incorporate the label information into $p(y|c, d_{\text{CFD}})$. We propose that information of each label exists in the hidden vectors of the inputs belonging to that label and denote such set of inputs for each label $l$ as $\mathcal{D}_{l}$. Inspired by the prototypical network \citep{snell2017prototypical}, we extract the information as follows:
\begin{equation}
\mathbf{h}^{l}_{[\text{CLS}]} = \frac{1}{|\mathcal{D}_l|} \sum_{S_j \in \mathcal{D}_{l}} \mathbf{h}_{j, [\text{CLS}]}.
\end{equation}
Hereafter, we forward both the topic-oriented representation $\mathbf{h}_{i}^{\text{topic}}$ and the label information in $\mathbf{h}^{l}_{[\text{CLS}]}$ to the non-linear layer to classify the counterfactuality as:
\begin{gather}
\mathbf{h’}_{[\text{CLS}]} = \text{Linear}\left(\left[\mathbf{h}^{\text{topic}}_{[\text{CLS}]}, \text{Linear}\left[\{\mathbf{h}^{l}_{[\text{CLS}]}\}_{l\in \mathcal{Y}}\right]\right]\right), \\
 p_{\text{CFD}} = p(y | \textup{do}(c)) = \frac{1}{|\mathcal{Y}|} \sum_{l} \varphi\left(\mathbf{h’}_{[\text{CLS}]}\right),
\end{gather}
where $\varphi$ denotes the sigmoid function.
\subsection{Training Strategy}
\label{subsect:training_strategy}
\noindent\textbf{Deconfounding NTM.} To deconfound the NTM, we derive the Eq. (\ref{eq:backdoor_adjustment}) as:
\begin{equation}
p(\boldsymbol{\theta} | \textup{do}(\mathbf{x}_{\text{bow}})) = \sum_{d_{\text{TM}}}
\frac{p(\boldsymbol{\theta}, \mathbf{x}_{\text{bow}} | d_{\text{TM}}) \cdot p(d_{\text{TM}})}{p(\mathbf{x}_{\text{bow}} | d_{\text{TM}})}.
\label{eq:initial_deconfounded_global_semantics}
\end{equation}
In NTMs, topics are parameterized as word distributions \cite{blum2016generalized, austin2011introduction}, similar to $\mathbf{x}_{\text{bow}}$. Hence, we conjecture that topic representation is a decomposed variant of each $\mathbf{x}_{\text{bow}}$, and we can only fully observe the distribution of the decompositions as in Figure \ref{fig:topic_distribution} with the same number of times we retrieve $\mathbf{x}_{\text{bow}}$. Furthermore, as the training progresses, the output $\mathbf{x’}_{\text{bow}}$ will converge to $\mathbf{x}_{\text{bow}}$. As such, we propose to approximate  Eq. (\ref{eq:initial_deconfounded_global_semantics}) following the propensity score modeling approach \citep{rosenbaum1983central}:
\begin{equation}
\begin{split}
p(\boldsymbol{\theta} | \textup{do}(\mathbf{x}_{\text{bow}})) \approx \sum_{d_{\text{TM}}} p(\boldsymbol{\theta}, \mathbf{x’}_{\text{bow}}, d_{\text{TM}})) \\
= \prod_{i} \frac{\boldsymbol{\phi}_{i} \cdot \boldsymbol{\theta}}{||\boldsymbol{\phi}_{i}|| \cdot ||\boldsymbol{\theta}||}, 
\end{split}
\label{eq:afterwards_deconfounded_global_semantics}
\end{equation}
where $i$ refers to a word in $\mathbf{x}_{\text{bow}}$, and we empirically add the magnitude of $\boldsymbol{\theta}$. The denominator works as a normalizer to balance the magnitude of the variables.

\noindent\textbf{Training Objective.} Our framework jointly optimizes the Neural Topic Model and Counterfactual Detection (CFD) module. To train the CFD module, we employ the binary cross-entropy loss as:
\begin{equation}
    \begin{split}
        &\mathcal{L}_{\text{CFD}} (S, \mathbf{x}_{\text{bow}}, y) = \\
        &-y \log p_{\text{CFD}} - (1-y) \log (1-p_{\text{CFD}}). 
    \end{split}
\end{equation}
For the NTM, with the Eq. (\ref{eq:afterwards_deconfounded_global_semantics}), we obtain the deconfounded evidence lower bound as:
\begin{equation}
\begin{split}
    &\mathcal{L}_{\text{NTM}}(\mathbf{x}_{\text{bow}}) = \text{KL}(q(\mathbf{z}|\mathbf{x}) || p(\mathbf{z})) \\ 
&- \mathbb{E}_{\epsilon \sim \mathcal{N}(\mathbf{0}, \mathbf{I})} \left[ \log p_{\boldsymbol{\phi}}(\mathbf{x}_{\text{bow}}|\boldsymbol{\theta}) \right] \\
&-\gamma \cdot \mathbb{E}_{\epsilon \sim \mathcal{N}(\mathbf{0}, \mathbf{I})} \left[\sum_{i=1}^{V} \log \frac{\boldsymbol{\phi}_{i} \cdot \boldsymbol{\theta}}{||\boldsymbol{\phi}_{i}|| \cdot ||\boldsymbol{\theta}||} \right],
\end{split}
\end{equation}        
where the first term denotes the Kullback-Leibler divergence between the prior and posterior distribution, the second term the reconstruction error of the output compared with the input, the third term the deconfounded objective in Eq. (\ref{eq:afterwards_deconfounded_global_semantics}), $V$ the vocabulary size, and $\gamma$ the hyperparameter to control the deconfounding effect upon the training, respectively. 

To conclude, our entire architecture is optimized with the linear combination of the loss functions $\mathcal{L}_{\text{NTM}}$ and $\mathcal{L}_{\text{CFD}}$ as: 
\begin{equation}
\mathcal{L} = \mathcal{L}_{\text{CFD}} + \lambda_{\text{NTM}} \mathcal{L}_{\text{NTM}},
\end{equation}
where $\lambda$ denotes the hyperparameter weight to scale the topic modeling component.
\section{Experiments}
\subsection{Datasets}
We evaluate on two prevalent datasets for the counterfactual detection task, SemEval-2020 \cite{yang2020semeval} and Amazon-2021 \cite{o2021wish}. While SemEval-2020 comprises English documents, Amazon-2021 covers statements in three languages, English, Japanese, and German. For our experiments, we inherit the original train/val/test splits. To verify the generalizability of our methods, we measure our performance on two other bias-sensitive document analysis tasks, Paraphrase Identification with the MRPC dataset \cite{dolan2005automatically}, and Implicit Sentiment Analysis (ISA) with CLIPEval from SemEval 2015 task 9 \cite{russo2015semeval}. These two tasks have been shown to sustain syntactic phrase and label biases \cite{li2020dice, wang2022causal}. The statistics of the datasets are provided in the Appendix. For evaluation metrics, we report Matthew’s Correlation Coefficient (MCC) \cite{boughorbel2017optimal}, the Accuracy (Acc), and F1 score.
\subsection{Implementation Details}
For the topic model, we select the topic number $T = 15$ based on the validation performance. Because at the beginning of the training process, the reconstructed output $\mathbf{x’}_{\text{bow}}$ does not resemble the input $\mathbf{x}_{\text{bow}}$, we decide to adapt the linear warm-up strategy \cite{gilmer2021loss} with the number of warm-up steps $N_{\text{wp}} = 1000$ for the value of $\gamma$ before fixing it to $0.25$. We finetune two pretrained multilingual language models, mBERT \citep{devlin2018bert} and XLM-R \citep{conneau2020unsupervised} for the CFD task, and the monolingual BERT \citep{devlin2018bert} and RoBERTa \citep{liu2019roberta} for the PI and ISA tasks. All variants are equipped with a linear layer on top of the pretrained language model. Our entire architecture is trained end-to-end on the A100 GPU, with the batch size of $16$ and $\lambda_{\text{NTM}}$ of $0.5$ for $50$ epochs, adopting Adam optimizer for the learning rate of $10^{-5}$ and L2 regularization equal $10^{-6}$. For the counterfactual detection and paraphrase identification tasks, $\mathcal{Y} = \{0, 1\}$, meanwhile for the implicit sentiment analysis task, $\mathcal{Y} = \{-1, 0, 1\}$.
\subsection{Baselines}
As baselines, we compare our work against a wide variety of recent state-of-the-art bias-resolving causal intervention and data augmentation approaches for Counterfactual Detection:
(i) \textbf{Stochastic Perturbation (SP)} \citep{wang2022causal}, leveraging word perturbation to causally intervene the spurious effect of the language bias confounder; (ii) \textbf{Masking} \citep{o2021wish}, masking clue phrases in counterfactual detection to eliminate their effect upon the training; (iii) \textbf{Debiased Focal Loss (DFL)} \citep{karimi2020end}, de-emphasizing the loss contribution of easy biased examples and direct the model towards hard but less biased ones; (iv) \textbf{Product of Experts} \citep{karimi2020end}, aggregating the predictions of two models, one trained with the biased and the other with both biased and unbiased examples; (v) \textbf{Backtranslation} \citep{o2021wish}, a data augmentation method on the input level to increase the number of rare-label samples.

\subsection{Comparison with State-of-the-arts}
\begin{table*}[h]
\centering
\resizebox{\linewidth}{!}{
\begin{tabular}{l|ccccccccc|ccc|ccc|ccc}
\hline
\multirow{3}{*}{\textbf{Methods}} & \multicolumn{9}{c|}{\textbf{Amazon-2021 (CD)}} & \multicolumn{3}{c|}{\textbf{SemEval-2020 (CD)}} & \multicolumn{3}{c|}{\textbf{MRPC (PI)}} & \multicolumn{3}{c}{\textbf{CLIPEval (ISA)}}\\ \cline{2-19}
& \multicolumn{3}{c}{\textbf{En}} & \multicolumn{3}{c}{\textbf{De}} & \multicolumn{3}{c|}{\textbf{Jp}} & \multicolumn{3}{c|}{\textbf{En}} & \multicolumn{3}{c|}{\textbf{En}} & \multicolumn{3}{c}{\textbf{En}} \\
 & Acc & MCC & F1 & Acc & MCC & F1 & Acc & MCC & F1 & Acc & MCC & F1 & Acc & MCC & F1 & Acc & MCC & F1\\
\hline
mBERT/BERT & 91.79 & 72.29 & 79.19 & 90.79 & 77.02 & 93.00 & 92.93 & 60.87 & 58.93 & 94.39 & 68.87 & 71.83 & 91.75 & 83.86 & 91.35 & 83.10 & 73.79 & 80.67 \\
w/ DFL & 93.88 & 81.70 & 81.30 & 91.11 & 79.58 & 93.47 & \underline{94.00} & \underline{66.61} & \underline{69.89} & \underline{96.63} & \underline{81.23} & \underline{82.80} & 92.15 & 84.62 & 91.69 & \underline{85.25} & \underline{77.22} & \underline{83.76} \\
w/ PoE & \underline{94.03} & 82.72 & 81.54 & 90.90 & 78.53 & 93.32 & 93.79 & 66.23 & 69.74 & 95.33 & 77.66 & 80.00 & 92.23 & 85.61 & 91.73 & 84.18 & 76.16 & 82.37 \\
w/ Backtranslation & \underline{94.03} & \underline{83.07} & \underline{81.89} & 90.26 & 73.51 & 92.47 & 93.25 & 62.50 & 60.27 & 95.94 & 79.18 & 80.99 & 92.30 & 85.95 & 91.75 & 83.10 & 74.22 &  82.33 \\
w/ Masking & 93.43 & 78.60 & 81.01 & 91.43 & 79.79 & 93.89 & 93.68 & 64.28 & 68.02 & 95.81 & 78.73 & 80.84 & --- & --- & --- & --- & --- & --- \\ 
w/ SP & 93.63 & 81.19 & 81.21 & \underline{91.65} & \underline{81.06} & \underline{93.86} & 93.61 & 62.76 & 66.83 & 95.08 & 77.35 & 79.02 & \underline{92.38} & \underline{86.53} & \underline{91.76} & 84.44 & 76.97 & 83.13 \\
\textbf{Our Model} & \textbf{95.52} & \textbf{86.37} & \textbf{83.05} & \textbf{92.29} & \textbf{82.08} & \textbf{94.40} & \textbf{95.29} & \textbf{73.79} & \textbf{75.00} & \textbf{96.97} & \textbf{83.31} & \textbf{84.81} & \textbf{93.65} & \textbf{87.38} & \textbf{93.20} & \textbf{86.79} & \textbf{79.28} & \textbf{85.00} \\ \hline
XLM-R/RoBERTa & 92.63 & 73.16 & 82.89 & 90.55 & 80.18 & 93.37 & 92.96 & 64.70 & 67.25 & 94.43 & 83.68 & 85.05 & 91.25 & 88.46 & 92.75 & 88.16 & 81.31 & 85.78 \\
w/ DFL & 94.66 & 85.22 & 83.46 & 90.84 & 80.32 & 93.53 & 94.40 & 74.21 & 75.68 & 96.41 & 84.43 & 85.56 & 93.92 & 88.99 & 94.05 & \underline{88.41} & 81.64 & 85.89 \\
w/ PoE & 94.52 & 84.90 & 83.09 & 90.86 & 80.58 & 93.74 & 94.12 & 73.23 & 75.59 & 96.43 & 84.55 & 85.87 & 93.50 & 88.91 & 93.75 & 87.87 & 80.80 & 84.99 \\
w/ Backtranslation & 94.95 & 85.85 & 83.81 & \underline{91.79} & \underline{81.02} & \underline{93.99} & 94.33 & 73.75 & 75.67 & 95.86 & 84.36 & 85.43 & 93.34 & 88.78 & 93.62 & 87.60 & 80.48 & 84.86 \\
w/ Masking & 94.20 & 83.75 & 82.96 & 90.21 & 79.75 & 93.25 & 94.57 & 74.44 & 75.73 & 95.79 & 84.22 & 85.11 & --- & --- & --- & --- & --- & --- \\
w/ SP & \underline{95.40} & \underline{86.35} & \underline{83.87} & 91.18 & 80.76 & 93.87 & \underline{94.87} & \underline{75.03} & \underline{76.85} & \underline{96.77} & \underline{85.18} & \underline{86.34} & \underline{94.60} & \underline{89.22} & \underline{94.17} & \underline{88.41} & \underline{81.69} & \underline{87.03} \\ 
\textbf{Our Model} & \textbf{96.85} & \textbf{88.74} & \textbf{84.44} & \textbf{92.51} & \textbf{82.49} & \textbf{94.57} & \textbf{95.82} & \textbf{76.01} & \textbf{77.97} & \textbf{97.44} & \textbf{86.09} & \textbf{87.44} & \textbf{95.55} & \textbf{91.05} & \textbf{95.14} & \textbf{89.49} & \textbf{83.55} & \textbf{88.07} \\ \hline
\end{tabular}}
\caption{Numerical results on original test sets of the Counterfactual Detection (CD), Paraphrase Identification (PI), and Implicit Sentiment Analysis (ISA) tasks. We respectively bold and underline the best and second-to-best results.}
\label{tab:exp_numerical_results}
\vspace{-15pt}
\end{table*}

\begin{table}[t]
\centering
\resizebox{0.7\linewidth}{!}{
\begin{tabular}{l|c|c|c}
\hline
\multirow{2}{*}{\textbf{Methods}} & \multicolumn{3}{c}{\textbf{SemEval-2020}}              \\
                                  & \multicolumn{1}{c}{Acc} & \multicolumn{1}{c}{MCC} & F1 \\ \hline
mBERT                             & \multicolumn{1}{c|}{85.75}    & \multicolumn{1}{c|}{74.21}    &  83.55  \\
w/ DFL                            & \multicolumn{1}{c|}{91.48}    & \multicolumn{1}{c|}{83.53}    &  90.94  \\
w/ PoE                            &          90.45               &       82.01                  &   89.60  \\
w/ Backtranslation                &              91.68           &        83.21                 &  91.54  \\
w/ Masking                        &           89.53              &       80.28                  &   88.52  \\
w/ SP                             &         90.60                &       81.92                  &   89.94  \\
Our model                         &            \textbf{93.13}             &         \textbf{86.52}                &  \textbf{92.84}  \\ \hline
XLM-R                             &          88.95               &        79.69                 &   88.04  \\
w/ DFL                            &          92.78               &     86.01                    &   92.38  \\
w/ PoE                            &      91.33                   &      83.50                   &   90.66 \\
w/ Backtranslation                &        92.48                 &     85.94                    &   92.49 \\
w/ Masking                        &        90.60                 &      82.15                   &  89.83  \\
w/ SP                             &          89.40              &      80.13                   &   88.25 \\
Our model                         &         \textbf{94.33}                &       \textbf{88.83}                  &  \textbf{94.14} \\ \hline
\end{tabular}}
\caption{Numerical results on balanced test sets of the CFD task on the SemEval-2020 dataset.}
\label{tab:balanced_experimental_results}
\end{table}

\noindent\textbf{Results on Original Test Sets.} We train and test the baselines and our model on the original test sets in Table \ref{tab:exp_numerical_results}. In the English variant of Amazon-2021 dataset, with mBERT we achieve an improved accuracy of $1.5$ points, and MCC score of $2.4$ points with XLM-R. For German documents, our XLM-R outperforms the baseline using Backtranslation with $1.5$ points, while our method adopted on mBERT enhances the SP approach with $1.0$ point in MCC. On the Japanese version, where the language upholds syntactic and morphological features separate from English and German, our mBERT-based and XLM-R-based models accomplish absolute enhancements of $5.1$ and $1.2$ points in F1 metric, respectively, compared with DFL and SP, which are the best previous approaches.

On the SemEval-2020 dataset, which is at a larger scale and concerns diverse domains \cite{o2021wish}, our general performance is also auspicious. In particular, our mBERT system surpasses the DFL model by a mean MCC of $2.1$ points. In addition, our XLM-R polishes the SP approach by $1.1$ points of the F1 score. These results corroborate that our counterfactual detection model is able to cope with harmful confounding impacts of different biases, thus producing more generalizable representations to attain better performance.

\noindent\textbf{Results on Balanced Test Sets.} We randomly sample 500 samples from each class in the test set of SemEval-2020, then evaluate our method in Table \ref{tab:balanced_experimental_results}. As can be seen, our method surpasses the best previous baseline, i.e. Backtranslation, with a significant margin of $1.5$ points of accuracy for the mBERT variant, and surpasses DFL with $2.8$ points of MCC for the XLM-R variant. These results verify that our causal intervention technique is able to mitigate the confounding effect of the class imbalance and makes the CFD model impartially consider the counterfactual and non-counterfactual choices when making prediction.

\subsection{Zero-Shot Cross-lingual Evaluation}
\begin{table}[t]
\centering
\small
\resizebox{\linewidth}{!}{
\begin{tabular}{ccccccc}
\hline
\multirow{2}{*}{\textbf{Models}} & \multicolumn{3}{c}{\textbf{Jp} $\rightarrow$ \textbf{En}} & \multicolumn{3}{c}{\textbf{De} $\rightarrow$ \textbf{Jp}} \\ 
& Acc & MCC & F1 & Acc & MCC & F1 \\ \hline
mBERT & 91.83 & 49.78 & 52.81 & 80.84 & 40.54 & 44.92 \\
\textbf{Our Model} & \textbf{93.40} & \textbf{59.41} & \textbf{60.71} & \textbf{91.54} & \textbf{50.10} & \textbf{54.34} \\
\hline
\end{tabular}}
\caption{
Cross-lingual Zero-shot mBERT results on the Amazon-2021 dataset.}
\label{tab:mbert_cross_lingual_results}
\vspace{-10pt}
\end{table}
\begin{table}[t]
\centering
\small
\resizebox{\linewidth}{!}{
\begin{tabular}{ccccccc}
\hline
\multirow{2}{*}{\textbf{Models}} & \multicolumn{3}{c}{\textbf{Jp} $\rightarrow$ \textbf{En}} & \multicolumn{3}{c}{\textbf{De} $\rightarrow$ \textbf{Jp}} \\ 
& Acc & MCC & F1 & Acc & MCC & F1 \\ \hline
XLM-R & 92.70 & 61.55 & 62.83 & 87.87 & 45.82 & 50.49 \\
\textbf{Our Model} & \textbf{93.85} & \textbf{62.79} & \textbf{64.35} & \textbf{89.19} & \textbf{55.56} & \textbf{51.67} \\
\hline
\end{tabular}}
\caption{
Cross-lingual Zero-shot XLM-R results on the Amazon-2021 dataset.}
\label{tab:xlmr_cross_lingual_results}
\vspace{-10pt}
\end{table}
To clearly confirm whether our methods have the ability to deal with the bias of clue phrases, we conduct the zero-shot cross-lingual evaluation. In particular, we proceed to finetune the standard mBERT, XLM-R, and our counterfactual detection architectures on the Japanese portion of Amazon-2021 dataset, then directly validate the performance on the English portion, and similarly we finetune the models on the German training set and test them on the Japanese test set. We indicate the results in Table \ref{tab:mbert_cross_lingual_results} and \ref{tab:xlmr_cross_lingual_results}.

As can be observed, our model is capable of enhancing zero-shot cross-lingual counterfactual detection capacity of both mBERT and XLM-R, surpassing mBERT with a large margin of $1.6$ and XLM-R with $1.1$ points in accuracy for the English test set. For the Japanese test set, we outperform mBERT and XLM-R with $9.6$ points of F1, and $9.7$ points of MCC, respectively. These results substantiate that our methods can mitigate the clue phrase bias in the language models.
\subsection{Adaptability to Other Bias-sensitive Tasks}
Experimental results in PI and ISA tasks are given in Table \ref{tab:exp_numerical_results}. For the MRPC dataset, our BERT model performance exceeds the one of the SP method by $1.3$ points in accuracy, and $1.4$ points in F1. With respect to the RoBERTA backbone, we also surpass the SP method by $1.8$ points of MCC, and $1.0$ point of F1. Regarding the CLIPEval dataset, integrating our approaches into BERT and RoBERTa extends the performance with $2.1$ points in MCC, and $1.1$ points in accuracy, respectively.

Those aforementioned results have shown that our methods have the capability of tackling biases in not only counterfactual detection but also other natural language understanding tasks.

\subsection{Ablation Study}
\begin{figure*}[t]
\centering
\includegraphics[width=0.8\linewidth] {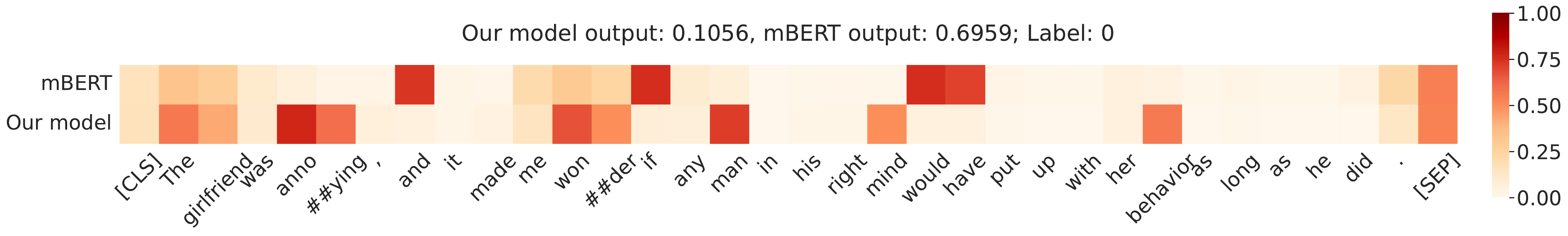}
\caption{Attention weights of the [CLS] token to all other words, and output scores of mBERT and Our Model. The score is in range $[0, 1]$. The input: ``\emph{The girlfriend was annoying, and it made me wonder if any man in his right mind would have put up with her behavior as long as he did.}''}
\label{fig:case_study}
\vspace{-10pt}
\end{figure*}
\begin{table}[t]
\centering
\small
\resizebox{0.9\linewidth}{!}{
\begin{tabular}{p{4cm}ccc}
\hline
\textbf{Methods} & \textbf{Acc} & \textbf{MCC} & \textbf{F1} \\ \hline
\textbf{Our Model} & \textbf{95.52} & \textbf{86.37} & \textbf{83.05} \\
w/o Debiased CFD Objective & 94.63 & 83.75 & 82.96 \\
w/o Deconfounded Topic Model Objective & 94.33 & 83.29 & 82.52 \\
w/o Neural Topic Model & 93.43 & 80.59 & 82.40 \\
\hline
\end{tabular}}
\caption{
Results from ablating different deconfounding components on the English Amazon-2021 dataset.}
\label{tab:ablation_components}
\vspace{-10pt}
\end{table}
\begin{table}[t]
\centering
\small
\resizebox{0.9\linewidth}{!}{
\begin{tabular}{p{4cm}ccc}
\hline
\textbf{Methods} & \textbf{Acc} & \textbf{MCC} & \textbf{F1} \\ \hline
XLM-R + NTM & \textbf{92.51} & \textbf{82.49} & \textbf{94.57} \\
XLM-R + PFA & 91.86 & 81.09 & 94.09 \\
XLM-R + LDA & 91.97 & 81.83 & 94.25 \\
\hline
\end{tabular}}
\caption{
Ablation results with various types of global semantics on the German Amazon-2021 dataset.}
\label{tab:ablation_global_semantics}
\vspace{-10pt}
\end{table}
\noindent\textbf{Effect of Deconfounding Components.} In this ablation, we experiment with removing each component that helps the model deconfound. Particularly, we train and test the ablated mBERT on the English portion of the Amazon-2021 dataset. As shown in Table \ref{tab:ablation_components}, solely employing one of the elements does enhance the overall counterfactual recognition, but being less effective than the joint approach. Without combining the deconfounding mechanisms, the model might not be able to cope with multiple biases.

\noindent\textbf{Effect of Global Semantics.} Here, we investigate the performance of our method when utilizing conventional topic models. We consider two choices, i.e. Poisson Factor Analysis (PFA) and Latent Dirichlet Allocation (LDA), and finetune the XML-R model on the German subset in Amazon-2021. As can be seen in Table \ref{tab:ablation_global_semantics}, NTM burnishes the counterfactual detector more effectively than traditional topic models. More results on the two ablation experiments can be found in the Appendix.

\subsection{Case Study}
\begin{figure}[t]
\centering
\includegraphics[width=0.8\linewidth] {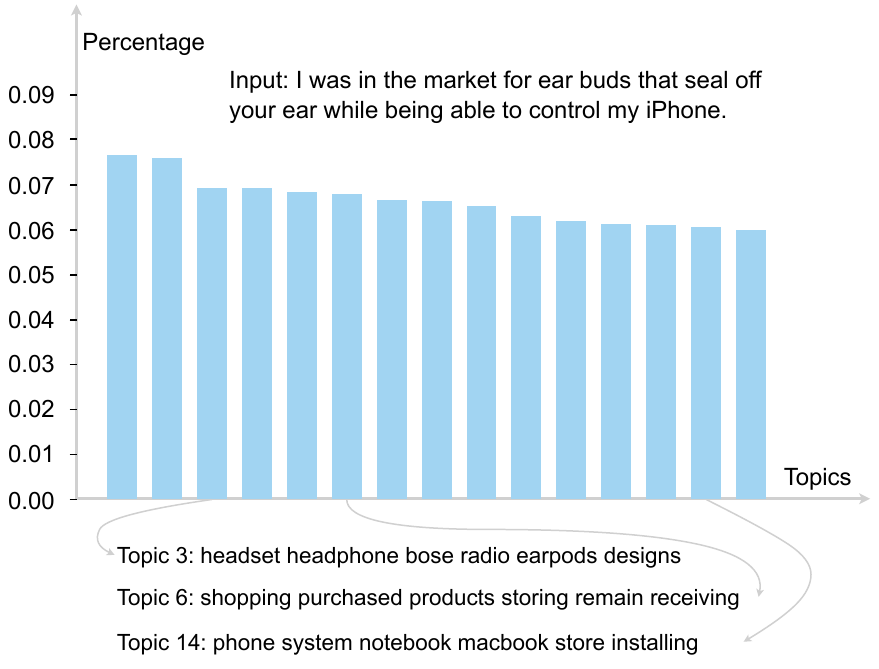}
\caption{Topic Percentages and inferred Top Topics from Figure \ref{fig:topic_distribution} after Causal Intervention.}
\label{fig:intervened_topic_distribution}
\vspace{-20pt}
\end{figure}

\noindent\textbf{Impact of Causal Intervention on Attention Logits.} Here we randomly select one example from Table \ref{tab:case_study} and visualize the average attention scores in the heads of all layers of the [CLS] token to the remaining words. As shown in Figure \ref{fig:case_study}, whereas mBERT’s [CLS] strongly pays attention to clue phrases``\emph{if}’’ and ``\emph{would have}’’, our model distributes the attention impartially and emphasizes content words, such as ``\emph{annoying}’’, ``\emph{man}’’, and ``\emph{behavior}’’. This could help to explain our more reasonable prediction than mBERT. We provide attention visualizations of other examples in the Appendix. These visualizations demonstrate that our approach can resolve the confounding influence of clue phrases and improve model prediction.

\noindent\textbf{Impact of Causal Intervention on Topic Distribution.} In Figure \ref{fig:intervened_topic_distribution}, we obtain topic representation from our neural topic model for the document of the Amazon-2021 dataset in Figure \ref{fig:topic_distribution}, and then count the percentage of documents sharing the top topic, i.e. possessing the largest likelihood. Different from Figure \ref{fig:topic_distribution}, our deconfounded topic model does not lean towards a subset of topics to assign top probabilities. Moreover, all three leading topics reveal the semantics of the document, which concerns \emph{headset}, \emph{shopping}, and \emph{phone}. These results demonstrate that our approach is capable of resolving the topic bias phenomenon to produce faithful global semantics for counterfactual detection.
\section{Conclusion}
In this paper, we propose a causal intervention framework that discovers biases in the counterfactual detection problem. In order to cope with clue phrase, topic, and label biases, we propose to utilize global semantics and extend the training strategy with deconfounding training objectives. Comprehensive experiments demonstrate that our model can ameliorate detrimental influences of biases to polish previous state-of-the-art baselines for not only the counterfactual detection but also other bias-sensitive NLU tasks. 
\section{Limitations}
We consider the following two limitations as our future work: (1) Extend the problem to circumstances with multiple observable confounding variables. The problem will become more complex if additional confounding factors are explicitly taken into account. Studying such complex scenario is potential to enhance the applicability and our understanding towards the proposed debiasing technique; (2) Explore the impact of causal intervention on generative tasks. We have only verified the effectiveness of causal intervention in discriminative language models. Whether the effectiveness applies for generative tasks such as machine translation, document summarization, etc., remains an open problem and interesting research direction.

\bibliography{anthology,custom}
\bibliographystyle{acl_natbib}
\clearpage
\appendix
\onecolumn
\section{Attention Visualization}
In this section, we visualize the attention weights of the [CLS] token to the words of the examples in Table \ref{tab:case_study}.

\begin{figure}[H]
\centering
\includegraphics[width=\linewidth] {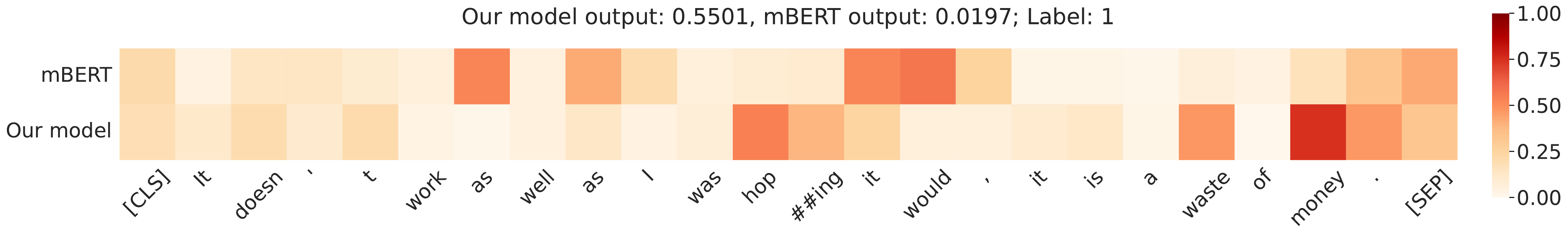}
\caption{Attention weights of the [CLS] token to all other words, and output scores of mBERT and our model. The score is in range $[0, 1]$. The input: ``\emph{It doesn’t work as well as I was hoping it would, it is a waste of money.}''}
\label{fig:app_case_study_1}
\end{figure}

\begin{figure}[H]
\centering
\includegraphics[width=\linewidth] {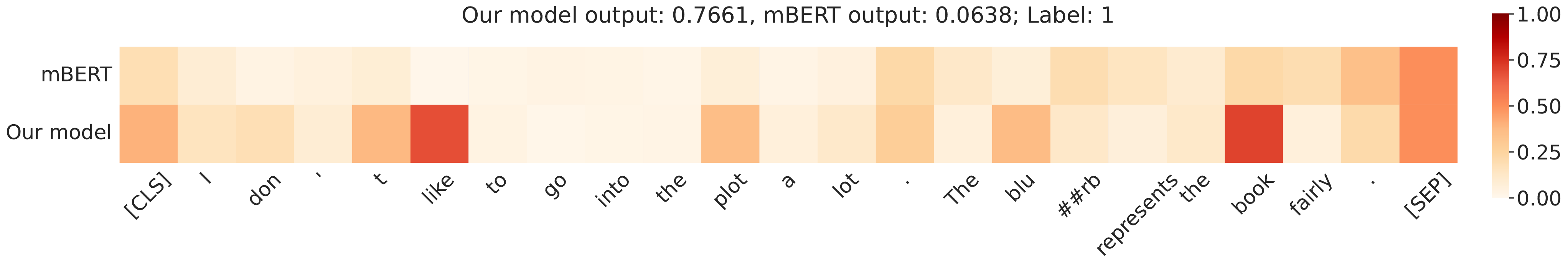}
\caption{Attention weights of the [CLS] token to all other words, and output scores of mBERT and our model. The score is in range $[0, 1]$. The input: ``\emph{I don’t like to go into the plot a lot. The blurb represents the book fairly.}''}
\label{fig:app_case_study_2}
\end{figure}

\begin{figure}[H]
\centering
\includegraphics[width=\linewidth] {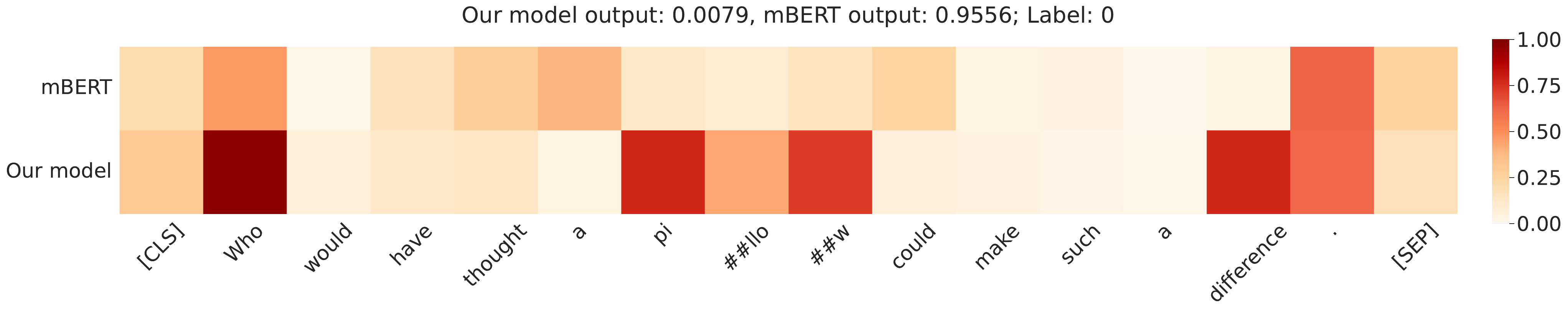}
\caption{Attention weights of the [CLS] token to all other words, and output scores of mBERT and our model. The score is in range $[0, 1]$. The input: ``\emph{Who would have thought a pillow could make such a difference.}''}
\label{fig:app_case_study_3}
\end{figure}

\begin{figure}[H]
\centering
\includegraphics[width=\linewidth] {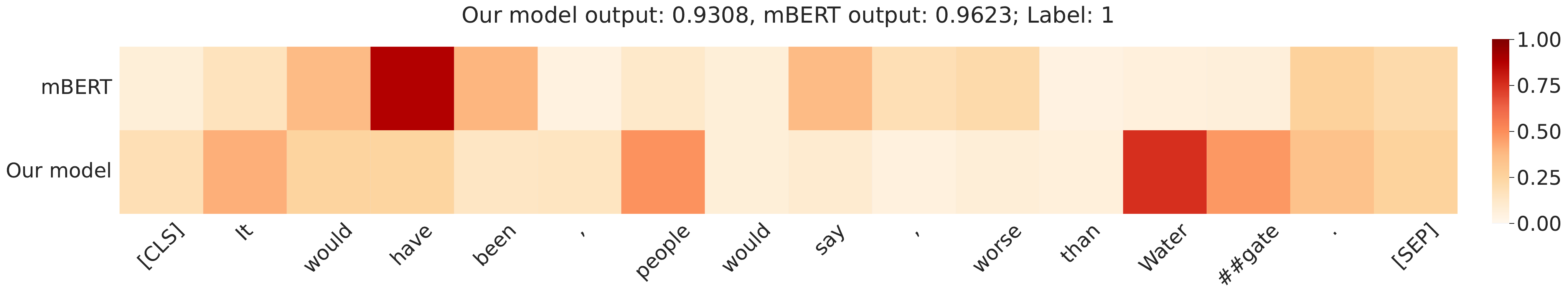}
\caption{Attention weights of the [CLS] token to all other words, and output scores of mBERT and our model. The score is in range $[0, 1]$. The input: ``\emph{It would have been, people would say, worse than Watergate.}''}
\label{fig:app_case_study_4}
\end{figure}

\begin{figure}[H]
\centering
\includegraphics[width=\linewidth] {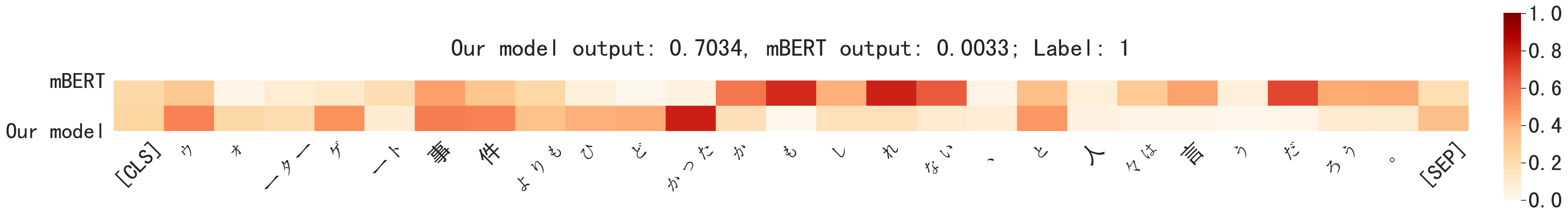}
\caption{Attention weights of the [CLS] token to all other words, and output scores of mBERT and our model. The score is in range $[0, 1]$. The input: \begin{CJK}{UTF8}{min}ウォーターゲート事件よりもひどかったかもしれない、と人々は言うだろう。\end{CJK}''}
\label{fig:app_case_study_5}
\end{figure}
    \section{Dataset Statistics}
In this section, we present the statistics of all the datasets pertaining to Counterfactual Detection, Paraphrase Identification, and Implicit Sentiment Analysis tasks.
\begin{table}[H]
\centering
\resizebox{\linewidth}{!}{
\begin{tabular}{ccccccccccc}
\hline
\textbf{\textbf{Dataset}}    & \textbf{\textbf{Variant}} & \textbf{\textbf{Train}} & \textbf{Val} & \textbf{Test} & \textbf{\#Pos} & \textbf{\#Neg} & \textbf{\#Neutral} & \textbf{\#Pos in Test} & \textbf{\#Neg in Test} & \textbf{Total} \\ \hline
\multirow{3}{*}{Amazon-2021} & En                        & 4018                    & 335          & 670           & 954            & 4069           & -                  & 131                    & 539                    & 5023           \\
                             & De                        & 5600                    & 466          & 934           & 4840           & 2160           & -                  & 650                    & 284                    & 7000           \\
                             & Jp                        & 5600                    & 466          & 934           & 667            & 6333           & -                  &     96                   &        838                & 7000           \\ \hline 
SemEval-2020                 & En                        & 13000                   & -            & 7000          & 2192           & 17808          & -                  & 738                    & 6262                   & 20000          \\ \hline 
MRPC                         & En                        & 49184                   & 2000         & 2000          & 23493          & 29691          & -                  &          907              &            1093            & 53184          \\ \hline 
CLIPEval                     & En                        & 1347                    & -            & 371           & 580            & 796            & 342                & 216                    & 155                    & 1718           \\ \hline 
\end{tabular}}
\caption{
Statistics of the Datasets.}
\label{tab:dataset_statistics}
\end{table}

\section{Additional Ablation Studies}

\textbf{Impact of Deconfounding Components.} We compare our model with its ablated variants in all subsets of the Amazon-2021 dataset. As can be observed in Table \ref{tab:exp_full_ablation_deconfounding_component}, jointly utilizing deconfounded neural topic model and debiased objective can tackle the clue phrase, label, and topic biases, leading to the largest overall improvement.
\begin{table*}[h]
\centering
\resizebox{0.95\linewidth}{!}{
\begin{tabular}{lccccccccc}
\hline
\multirow{2}{*}{\textbf{Methods}} & \multicolumn{3}{c}{\textbf{En}} & \multicolumn{3}{c}{\textbf{De}} & \multicolumn{3}{c}{\textbf{Jp}} \\
 & Acc & MCC & F1 & Acc & MCC & F1 & Acc & MCC & F1 \\
\hline
Our Model & \textbf{95.52} & \textbf{86.37} & \textbf{83.05} & \textbf{92.29} & \textbf{82.08} & \textbf{94.40} & \textbf{95.29} & \textbf{73.79} & \textbf{75.00} \\
w/o Debiased CFD objective & 94.63 & 83.75 & 82.96 & 92.15 & 81.49 & 94.36 & 94.87 & 73.13 & 73.61 \\
w/o Deconfounded Topic Model objective & 94.33 & 83.29 & 82.52 & 91.94 & 81.31 & 94.19 & 94.80 & 72.78 & 71.66 \\
w/o Neural Topic Model & 93.43 & 80.59 & 82.40 & 91.76 & 81.20 & 94.05 & 94.72 & 72.60 & 71.09 \\ \hline
\end{tabular}}
\caption{Results of subsequently pruning deconfounding components on the Amazon-2021 dataset.}
\label{tab:exp_full_ablation_deconfounding_component}
\end{table*}

\noindent\textbf{Impact of Global Semantics.} In addition to the results in Table \ref{tab:ablation_global_semantics}, we execute our model with different topic models on other languages of the Amazon-2021 dataset. The results are shown in Table \ref{tab:exp_full_ablation_global_semantics}.

\begin{table*}[h]
\centering
\resizebox{0.9\linewidth}{!}{
\begin{tabular}{lccccccccc}
\hline
\multirow{2}{*}{\textbf{Methods}} & \multicolumn{3}{c}{\textbf{En}} & \multicolumn{3}{c}{\textbf{De}} & \multicolumn{3}{c}{\textbf{Jp}} \\
 & Acc & MCC & F1 & Acc & MCC & F1 & Acc & MCC & F1 \\
\hline
XLM-R + NTM & \textbf{96.85} & \textbf{88.74} & \textbf{84.44} & \textbf{92.51} & \textbf{82.49} & \textbf{94.57} & \textbf{95.82} & \textbf{76.01} & \textbf{77.97} \\
XLM-R + PFA & 96.63 & 87.19 & 83.97 & 91.86 & 81.09 & 94.09 & 95.72 & 75.46 & 77.68 \\
XLM-R + LDA & 96.40 & 87.11 & 83.10 & 91.97 & 81.83 & 94.09 & 95.18 & 75.07 & 77.53 \\ \hline
\end{tabular}}
\caption{Results with heterogeneous topic models on the Amazon-2021 dataset.}
\label{tab:exp_full_ablation_global_semantics}
\end{table*}

\end{document}